\documentclass[letterpaper, 10 pt, conference]{ieeeconf}

\IEEEoverridecommandlockouts
\usepackage{graphics} % for pdf, bitmapped graphics files
\usepackage{epsfig} % for postscript graphics files
\usepackage{subfigure}
\usepackage{textcomp}
\usepackage{gensymb}
\usepackage[space]{cite}

\usepackage{mathptmx} % assumes new font selection scheme installed
\usepackage{times} % assumes new font selection scheme installed
\usepackage{amsmath} % assumes amsmath package installed
\usepackage{amssymb}  % assumes amsmath package installed
\usepackage[ruled,linesnumbered]{algorithm2e}
\usepackage{multirow}

\usepackage{cite}
\usepackage[colorlinks=true,linkcolor=black,urlcolor=blue,citecolor=black]{hyperref}

\usepackage{algorithmic}
\begin{document}

\title{Contact Pose Identification for Peg-in-Hole Assembly under Uncertainties
}

\author{Shiyu Jin, Xinghao Zhu, Changhao Wang, and Masayoshi Tomizuka% <-this % stops a space %
% \thanks{*: Both authors contributed equally to this work.}% <-this % stops a space
\thanks{Department of Mechanical Engineering, University of California, Berkeley, CA, USA.
        {\tt\small \{jsy, zhuxh, changhaowang, tomizuka\}@berkeley.edu}}%
}

%\author{\IEEEauthorblockN{1\textsuperscript{st} Given Name Surname}
%\IEEEauthorblockA{\textit{dept. name of organization (of Aff.)} \\
%\textit{name of organization (of Aff.)}\\
%City, Country \\
%email address}
%\and
%\IEEEauthorblockN{2\textsuperscript{nd} Given Name Surname}
%\IEEEauthorblockA{\textit{dept. name of organization (of Aff.)} \\
%\textit{name of organization (of Aff.)}\\
%City, Country \\
%email address}
%\and
%\IEEEauthorblockN{3\textsuperscript{rd} Given Name Surname}
%\IEEEauthorblockA{\textit{dept. name of organization (of Aff.)} \\
%\textit{name of organization (of Aff.)}\\
%City, Country \\
%email address}

%}

\maketitle

\begin{abstract}
Peg-in-hole assembly is a challenging contact-rich manipulation task. There is no general solution to identify the relative position and orientation between the peg and the hole. In this paper, we propose a novel method to classify the contact poses based on a sequence of contact measurements. When the peg contacts the hole with pose uncertainties, a tilt-then-rotate strategy is applied, and the contacts are measured as a group of patterns to encode the contact pose. A convolutional neural network (CNN) is trained to classify the contact poses according to the patterns. In the end, an admittance controller guides the peg towards the error direction and finishes the peg-in-hole assembly. Simulations and experiments are provided to show that the proposed method can be applied to the peg-in-hole assembly of different geometries. We also demonstrate the ability to alleviate the sim-to-real gap.

\end{abstract}

%%%%%%%%%%
\section{Introduction}

Robotic peg-in-hole assembly has been studied for decades. It is challenging because it requires accurate state estimations of the peg and the hole for alignment, and a combination of precise planning and control algorithms for insertion. 

Identifying the contact pose, the relative position and orientation between the peg and the hole, is required to align the peg and the hole before insertion. Visual feedback is the most common strategy to identify the pose \cite{pose_estimate1,pose_estimate2}. However, vision sensors suffer from high precision requirements and occlusions during the assembly task. In order to avoid such problems, search-based algorithms such as random search or spiral search \cite{2001search} have been proposed to compensate the uncertainties of contact pose. The search strategy generates a search path within the search area for hole localization, which is not efficient especially when the search area is large and the search dimension is high. % I do not know what the search method is. Not very clear here

For insertion, %In the insertion phase, the small clearance makes the problem hard. 
the clearance between the peg and the hole is usually smaller than the precision of a robot. A tiny position and orientation error could cause workpieces to jam and wedge and may lead to failure or even damage to the workpieces. Compliance, either passive or active, has shown to be effective in handling the small uncertainties of position and orientation. Passive compliance utilizes passive compliance hardwares such as RCC \cite{Drake1978UsingCI,Whitney} to compensate uncertainties. In constrast, active compliance applies control strategies from software to let the robot mimic the spring-damping behavior \cite{handbook,admittance}.
%utilizes force control such as impedance control or admittance control \cite{handbook,admittance}. The general idea is to make the system compliant to the environment and minimize the contact force during assembly.

%When the position and orientation error are large, insertion cannot be conducted even with the existence of compliance. We define the contact pose as the relative position and orientation between the peg and the hole. An alignment procedure is required to identify the contact pose before insertion. Computer vision is one solution for identifying the pose of the assembly parts. But vision sensors suffer in peg-in-hole assembly tasks when there are calibration errors or visual information of the hole is not available due to the occlusion of the peg. Random search or spiral search strategy [cite?] is another solution to compensate contact pose uncertainties without vision system. Search strategy generates a search path within the search area, which is not efficient especially when the search area is large and search dimension increases.

\begin{figure}
	\centering
	\includegraphics[scale=0.25]{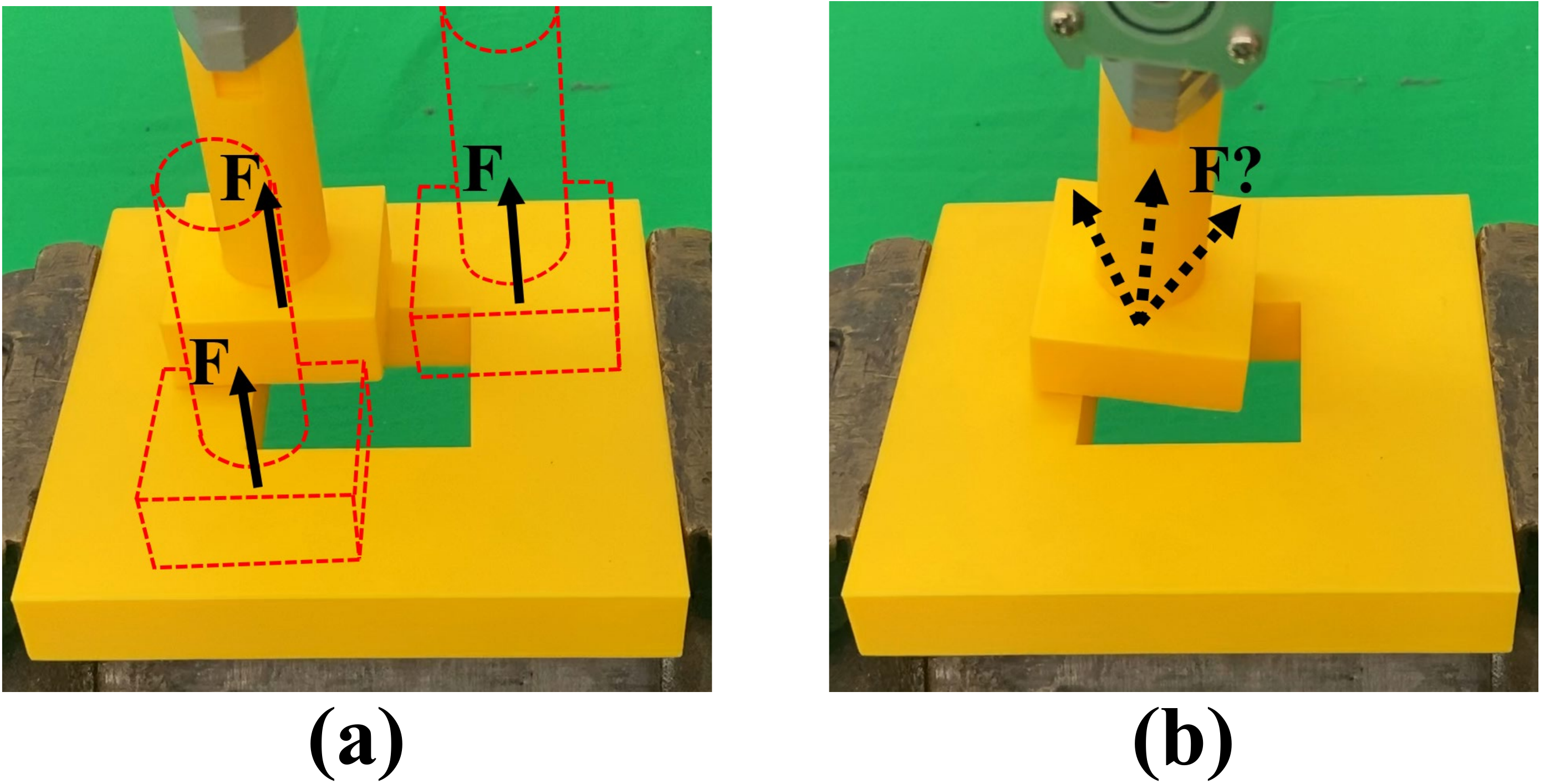}
	\caption{(a) The same upward contact force could come from many possible contact poses. (b) The same contact pose could generate many possible contact forces within the Coulomb friction cone.}
	\label{fig:contact_pose}
\end{figure}

% To address the limitations of vision system and search strategy, force/torque sensor has been widely used in assembly tasks.
In contact-rich scenarios, force/torque-based method normally conveys more information than vision-based and search-based methods. Tang\cite{2016threepoint} analyzed a three-point contact model for round peg and hole. But the method lacked the ability to generalize to complex geometries. Kim proposed a peg shape recognition and hole detection algorithm using the force/torque sensor by inclining the peg in all directions, but their method suffered from the cumulative error\cite{2014rotate}. In recent years, many learning-based methods have been proposed to solve the peg-in-hole assembly problem\cite{2016teaching,2016endtoend,2019yongxiang,2019syntheticdata,2019makingsense}. They
%Comparing with vision based and search based strategies, n in this contact-rich scenario. 
%Force/torque feedback provides observations of current contact conditions between object and environment for accurate localization and control\cite{2019makingsense}. 
treated the task as a Markov decision process, where the contact feedback at the current time step is used to determine the action of the next step. However, the mapping from the force/torque feedback to the contact pose is not injective as shwon in Fig. \ref{fig:contact_pose}. % injective?
On one hand, the same contact forces can be measured at different contact poses. On the other hand, the same contact pose could generate different contact forces, i.e. all the possible forces within the Coulomb friction cone. To deal with the above problem, particle filter was applied to identify the location based on multiple observations in \cite{2005particle,2007particle}. However, it is time-consuming to generate the force-position mapping in the real world and hard to generalize. %Tang[5] proposed  an autonomous alignment method based on a three-point contact model to align peg and hole before insertion. But the proposed method only works for round peg-hole.
%xxx. These methods, however, are time-consuming

% 你要不先把这些model based 的方法放在MDP   前面讲？
%Tang\cite{2016threepoint} analyzed a three-point contact model for round peg and hole. But the method lacked the ability to generalize to complex geometries. Kim proposed a peg shape recognition and hole detection algorithm using the force/torque sensor by inclining the peg in all directions. But their method suffered from cumulative error \cite{2014rotate}.
%A follow-up work studied on the assembly direction selection method for complex-shaped parts using the geometric information \cite{2014cad}. However, it requires an accurate vision sensor.

In this paper, we propose a novel method that can identify the contact poses based on a sequence of contact measurements. At initialization, the peg contacts the hole with pose uncertainties. The peg then follows a designed tilt-then-rotate motion to make contact with the hole. % to collect contact measurements. The sequence of measurements are then projected to image plane. An injective mapping between the image pattern and contact poses is learned by a CNN. The CNN predicts the contact poses based on the error directions. Given the contact poses, an admi ...
The contact measurements are plotted in polar coordinates to generate a group of patterns. An injective mapping between the patterns and contact poses is learned by a convolutional neural network (CNN), % An injectie mapping between the patterns and contact poses is learned by a CNN?
which classifies the contact poses based on the error directions. Finally, an admittance controller will guide the peg towards the error direction and finish insertion. There are two main contributions of this paper. 1) We construct the mapping using a sequence of measurements as input instead of feedback at one single time step. This makes the mapping become one-to-one. 2) We classify the contact pose based on patterns, which improves the generalization ability of the proposed method. It can even tackle the sim-to-real gap.

% There are several advantages of the proposed method. % You should mention why.
% The mapping from pattern to contact pose is a one-to-one mapping. 2)The pattern is robust to sensor noise, because the pattern is constructed using a sequence of contact measurement instead of one single measurement. 3)This contact pose identification model is easy to obtain, because all the training data can be generated in simulation in a very short time. 4)This model has good generalization. Since the measurement data is normalized and plotted in a polar coordinate, the pattern is sensitive neither to the size nor the material of the parts. A model learned from a smaller peg and hole can be successfully applied to a larger one as long as their shapes are the same. What’s more, the model learned in simulation can be successfully adapted to real-world.%, even although the physical properties have huge differences, such as the size, material, friction factor, and many mechanic constraints in the real robot.

% Since the pattern is plotted using normalized data, the pattern classifier learned in simulation can be directly applied to real-world peg-in-hole tasks for peg-hole of different sizes and materials as long as the peg-hole shapes are identical. 

The remainder of this paper is organized as follows. Section II introduces the background including task description, admittance control, and assembly strategy. Section III describes the proposed contact pose identification method according to contact patterns. Section IV shows the performance of the proposed method by both simulations and real-world experiments. Section V discusses the advantages and disadvantages of the proposed method and proposes future work.

\section{Background}

\subsection{Task Description}
% we focus on ... Generally speaking, the pose of the peg and the hole might be noisy due to sensor inaccuracy. To simply the problem, we assume the pose of the peg is static during the assembly, which can be computed using the robotic forward kinematics. The uncertainty are introduced by the hole's pose.
We focus on the peg-in-hole assembly task under pose uncertainties. 
%The peg is fixed with the robot end-effector, and the pose is obtained via forward kinematics. The hole is fixed on the table, and the pose can be estimated by a vision system with uncertainties in 6 degrees of freedom (DOF). 
Generally speaking, the pose of the peg and the hole might be noisy due to sensor inaccuracy. To simply the problem, we assume the peg is fixed with the robot end-effector, and the pose can be obtained via forward kinematics. The hole is fixed on the table, and the pose can be estimated by a vision system with uncertainties in 6 degrees of freedom (DOF). The magnitudes of the uncertainties are roughly $\pm 20 mm$ and $\pm 3 \degree$ for position and orientation respectively, which are determined by the precision of the visual system. The clearance between the peg and the hole is $1 mm$. %We define the contact pose as the relative pose between the peg and the hole. 

% achieve the peg-in-hole assembly?
The goal of the task is to compensate the uncertainties of contact pose and achieve the peg-in-hole assembly. The contact surfaces of both the peg and the hole are assumed to be flat. %We also assume the knowledge of the CAD model of the assembly parts.

\subsection{Admittance Control}

Admittance control \cite{handbook,admittance} is widely used in robotic manipulation tasks to handle contact dynamics. By adding a virtual spring-damping system, the contact between the robot and the environment becomes soft, which improves the manipulation performance and prevents from damaging either the robot or the environment. We apply admittance control to the following assembly strategy to track the desired peg trajectory and compensate small uncertainties in assembly.

\begin{figure}
	\centering
	\includegraphics[scale=0.25]{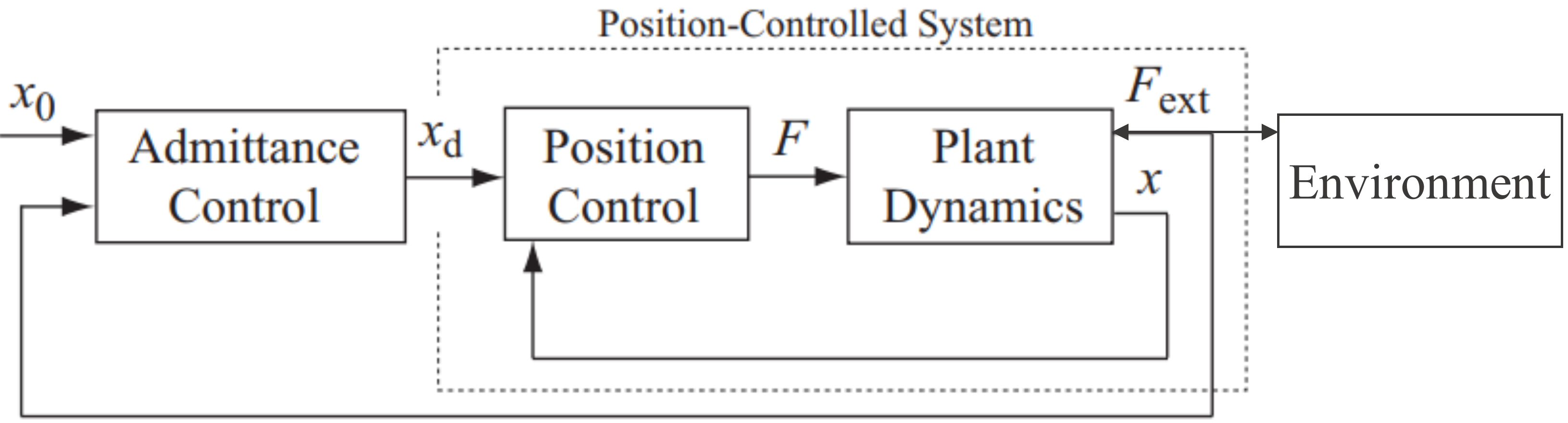}
	\caption{Admittance Control.}
	\label{fig:blocks}
\end{figure}

In admittance control, the desired pose $x_0$ and measured external force/torque $F_{ext}$ are inputs to the admittance control block (Fig. \ref{fig:blocks}), which generates the reference pose $x_d$ for the PD position control.

\begin{align}
F + F_{ext} & =m\ddot{x} \\
F & = k_p(x_d - x) - k_d \dot{x}\\
 F_{ext} & = M_d(\ddot{x}_d -\ddot{x}_0) + D_d(\dot{x}_d - \dot{x}_0) +K_d(x_d - x_0)
\label{eqn_1}
\end{align}
where $M_d$, $D_d$, and
$K_d$ represent the desired inertia , damping, and stiffness, respectively. $k_p$ and $k_d$ are PD position control gains.

\subsection{Assembly Strategy}
\label{assembly_strategy}

Peg-in-hole assembly has been studied for decades. An efficient and widely used assembly strategy divides the task into several stages \cite{2014rotate,para_tunning}: initialization, approaching, contact pose estimation,  alignment, and insertion. At initialization, the peg and the hole are fixed on the robot manipulator and the table, respectively. A vision system is applied to roughly estimate the pose of the hole. At approaching, the peg approaches to the hole % with admittance control?
with an admittance controller. With well-tuned controller parameters, the plane contact between the flat surface of the peg and the hole could eliminate the pose uncertainties in 3 dimensions, roll axis, pitch axis, and z-axis. At contact pose estimation, the peg explores along the surface of the hole to estimate the relative position and orientation between the peg and the hole. This stage eliminates the uncertainties in x and y axes. Finally, based on the contact pose estimation, the peg can slide towards the hole and finish insertion with an admittance controller. Small oscillation is added to the yaw axis in this stage, together with admittance control, to compensate small uncertainties of yaw axis. In this paper, we mainly focus on the contact pose estimation stage, which is introduced in section \ref{proposed_method}.

%%%%%%%%%%%%%%%
\section{Proposed Method}
\label{proposed_method}

%Most of the assembly strategies, such as spiral search, force feedback control, or reinforcement learning, treat the assembly task as a Markov decision process. They use the contact feedback state at one single time step to make the assembly decision for the next step. However, the mapping from force/torque feedback to peg-hole contact pose is not a one-to-one mapping (Fig. \ref{fig:contact_pose}). This brings difficulties to identify contact pose according to force/torque feedback. %Hence, (the assembly strategy based on feedback force/torque at one single time step is not informative.)?
\begin{figure*}[!ht]
	\centering
	\includegraphics[scale=0.53]{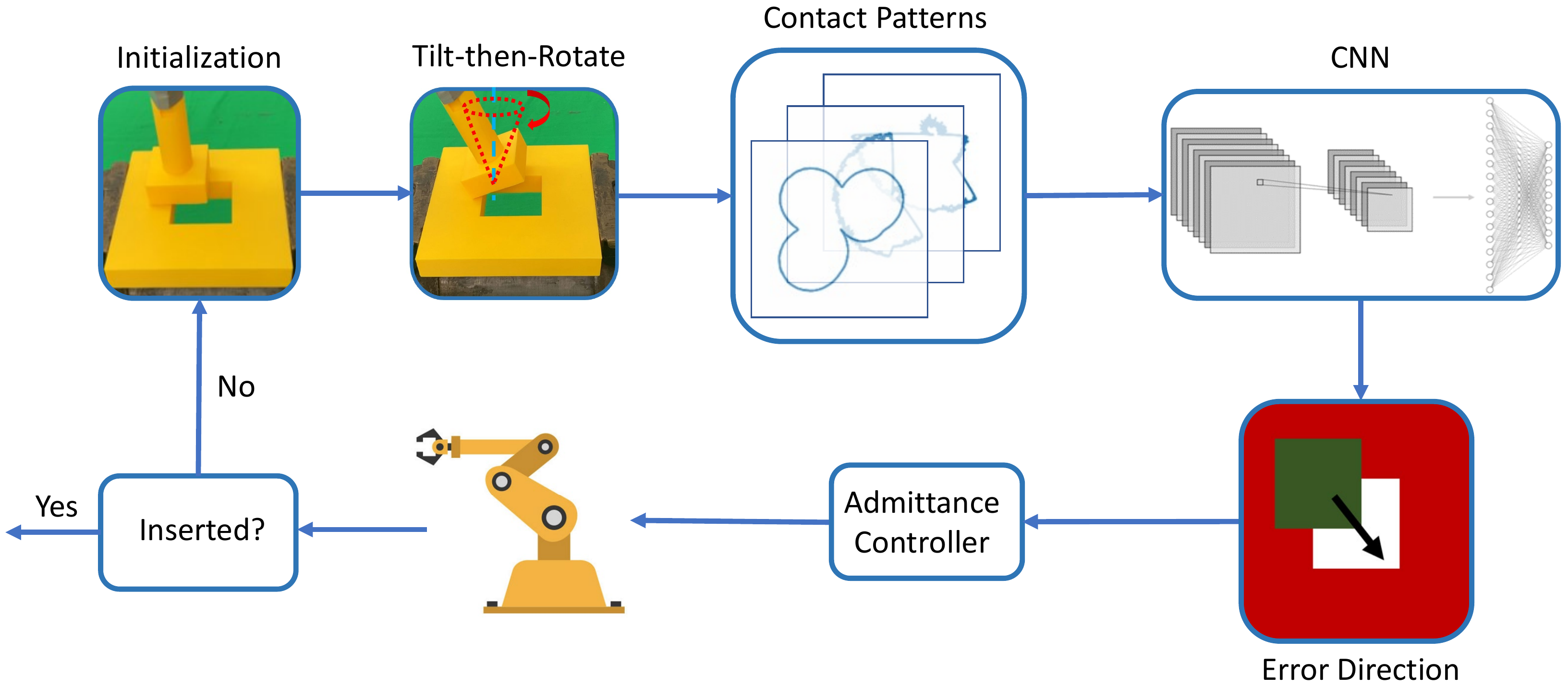}
	\caption{Framework of the proposed method.}
	\label{fig:framework}
\end{figure*}

Peg-in-hole assembly can be accomplished easily by a human even with eyes closed.
%Considering human doing assembly tasks, a human holds the peg in hand and tries to identify the hole location with eyes closed. % no "make initial"?
The human will first use the peg to make contact with the hole. Then he/she will locally move the peg to sense hole's location based on a sequence of contacts instead of just one single contact. If there is a hole in one direction, the tip of the  peg could slide into the hole a little bit and the force/torque feedback also have an impulse in that direction. %??? 
Based on the historical measurements in a sequence of contacts, the human keeps updating the knowledge of the contact pose and eliminating the hole uncertainties. 

Inspired by the human strategy, we propose to use a sequence of contact feedback to identify the contact pose under uncertainties (Fig. \ref{fig:framework}). 

\subsection{Tilt-then-Rotate Strategy}

\begin{figure*}[!ht]
	\centering
	\includegraphics[scale=0.54]{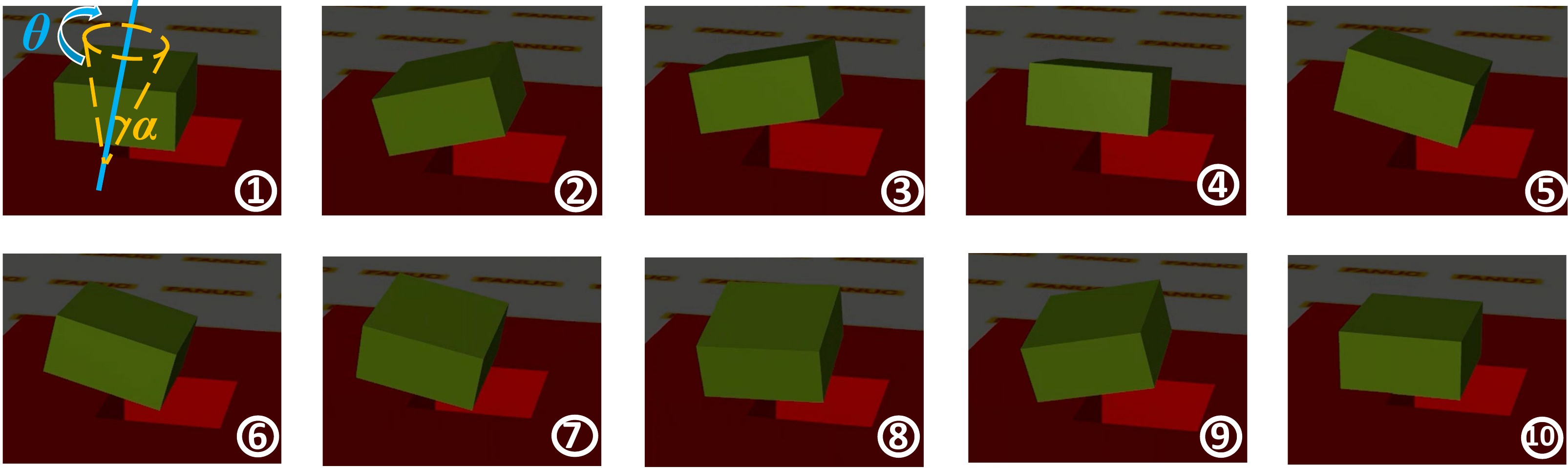}
	\caption{Snapshots of the tilt-then-rotate strategy. The blue line is z-axis. The yellow cone represents the designed trajectory for rotation. Tilt (2) then rotate (2-9) the peg for $2\pi$. While the peg is being rotated, a constant downward force is applied to maintain a single point contact (3,5,9), line contact (2,4,6,8), or two points contact (7) between the peg and the hole.}
	\label{fig:tilt_then_rotate}
\end{figure*}

The peg contacts the hole after the approaching stage (Fig. \ref{fig:tilt_then_rotate}.1). The peg and the hole have some overlaps but are not aligned well due to the uncertainties of the contact pose. We propose a tilt-then-rotate strategy to identify the contact pose. 

% we parametrized the rotation-and -tilt trajeectory using alpha and theta as: ..., in which alpha is ... theta is ... To be more specific, the peg will tilt for alpha degrees in ...
We tilt the peg for $\alpha$ degrees in all directions by rotating the peg for $2 \pi$ (Fig. \ref{fig:tilt_then_rotate}). The tilt-then-rotate trajectory can be described as continuously changing $\theta$ from $0$ to $2 \pi$ in order to change the roll and the pitch angle:
\begin{equation}
    \{roll, pitch\} = \{\alpha sin(\theta), \alpha cos(\theta) \}, \quad  \theta \in [0,2 \pi)
\label{roll_pitch}
\end{equation}
The desired tilt-then-rotate trajectory is tracked by an admittance controller. At the same time, a constant downward force is applied to the peg in order to  % no always?
maintain contact with the hole. During the procedure, contact force and torque are measured by a force/torque sensor. % mounted on the robot.
As the peg is tilted in all directions, the contact keeps switching between one point contact, two points contact, and line contact (Fig. \ref{fig:tilt_then_rotate}.2-\ref{fig:tilt_then_rotate}.9). The tip of the peg could go into the hole when the peg tilts towards the hole and the force/torque measurements would also have an impulse. Different contact poses will result in different sequences of measurements along the designed tilt-then-rotate trajectory. Comparing with one measurement at a single time step, the mapping from a sequence of measurements to contact poses becomes an injective mapping.

\subsection{Contact Pattern Generation}
% what 12 dim?
The tilt-then-rotate strategy generates a sequence of measurements in 12 dimensions including force ($\mathbb{R}^3$), torque ($\mathbb{R}^3$), and peg pose ($\mathbb{R}^6$). For different control forces or different sizes of the parts, those measurements can be different in the order of magnitude. % no in contrast ?
Human can sense the contact pose in different scenarios by the same exploring strategy. There must be some high-level features we can extract from the measurements. % this looks weird

\begin{figure}
	\centering
	\includegraphics[scale=0.35]{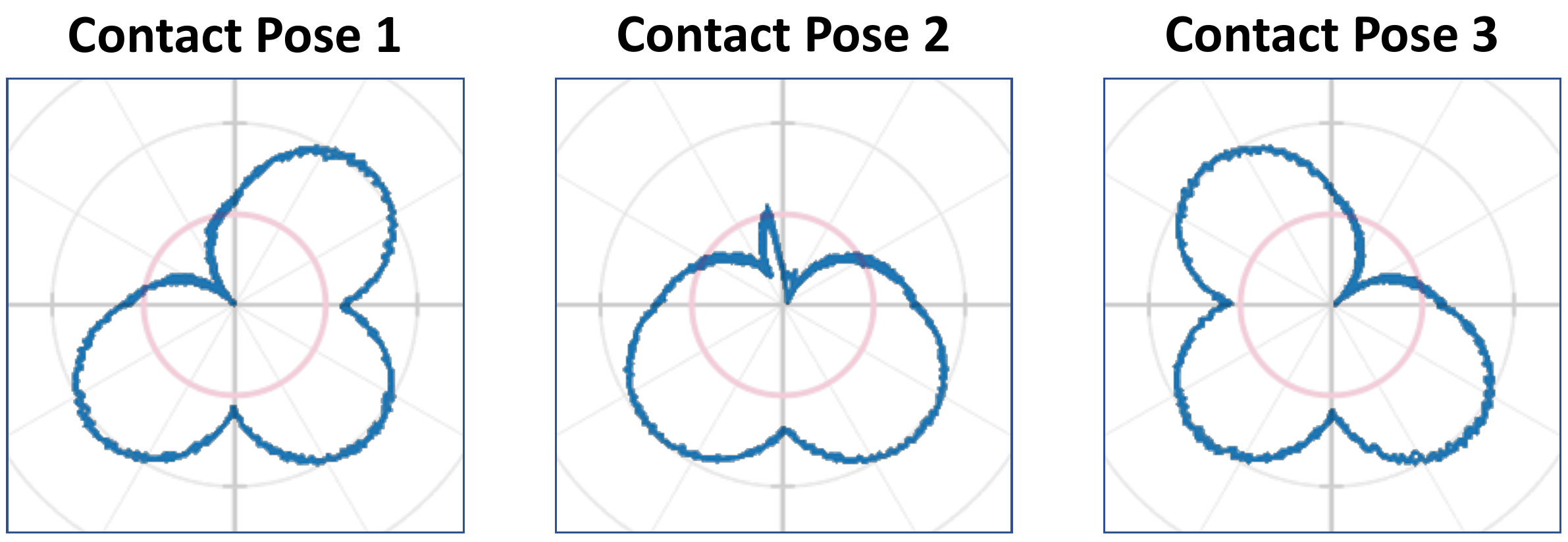}
	\caption{Contact patterns in polar coordinates for 3 different contact poses. Only z-axis channel is shown.}
	\label{fig:contact_pattern}
\end{figure}

We propose to plot the measurements of each dimension in polar coordinate as one channel. % ??? We propose to project the contact measurements in each channel to polar coordinate, results in 12 images ...
The data in each channel is normalized, then smoothed by moving average. The normalization makes the data invariant to control forces and sizes of the parts. The moving average reduces the sensor noises. % We utilized these 12 channels of polar coordinate data as one contact pattern, ...
We utilize the plotted image with 12 channels as one contact pattern, which encodes high-level features about the contact pose. Fig. \ref{fig:contact_pattern} shows z-axis channel of the contact pattern for different contact poses.

% 先讲cnn再说这个？
One contact pose corresponds to one contact pattern with 12 channels. In order to construct an informative mapping, we need to perform hundreds of tilt-then-rotate motions for all contact poses. This is not only time-consuming but also inaccurate due to the limitation of pose sensing in the real world. We propose to generate the contact pattern in the MuJoCo physics engine. The simulated environment can perform hundreds of trails in a short time. In addition, ground truth contact pose can be obtained easily in simulation (Fig. \ref{fig:tilt_then_rotate}).

\subsection{Contact Pose Classification Neural Network}

\begin{figure}
	\centering
	\includegraphics[scale=0.5]{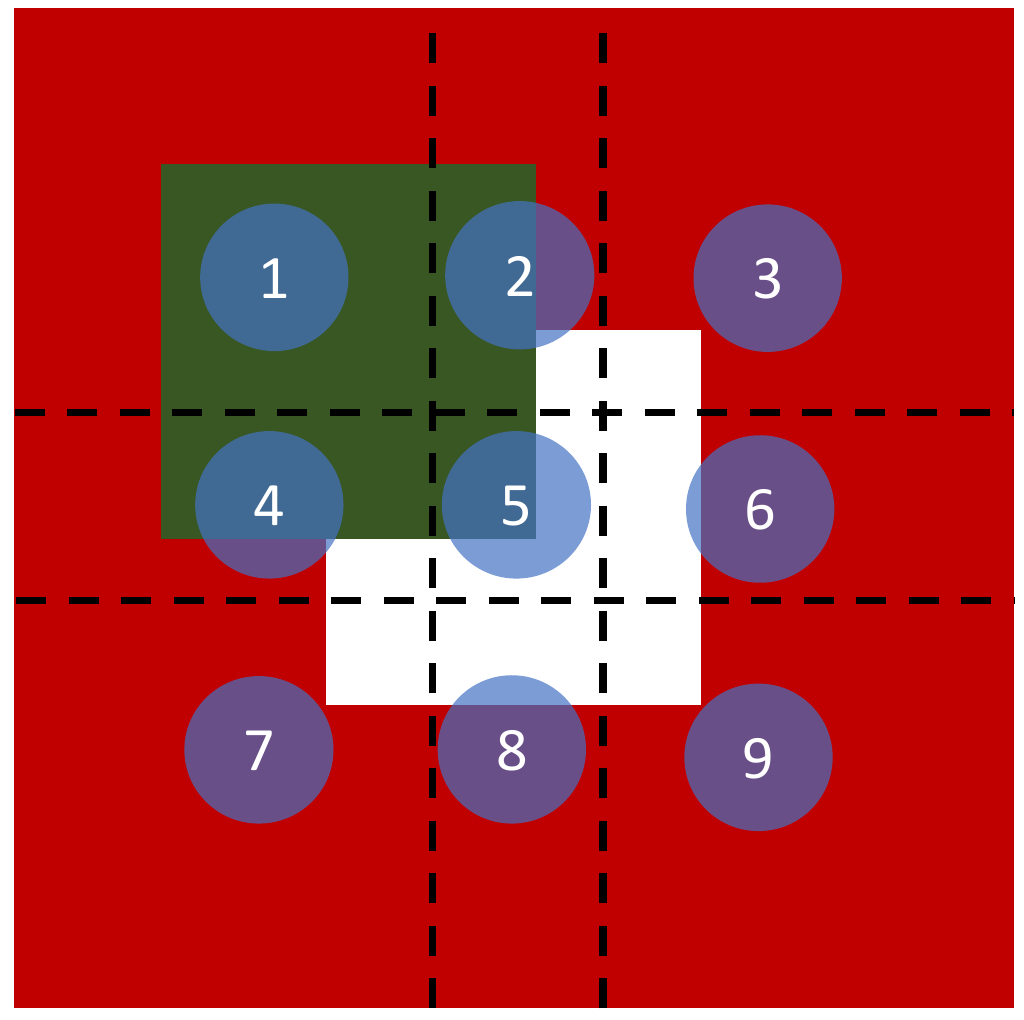}
	\caption{The contact poses are classified into 9 classes according to which edge of the peg contacts the hole.}
	\label{fig:classes}
\end{figure}

% Given contact pattern, a CNN was used to predict the contact states.
% We treated the ... as an image recognition problem.
% which 3 channels?
Contact poses of a square peg-hole can be classified into 9 classes according to which edge of the peg contacts the hole (Fig. \ref{fig:classes}). Each class of contact pose has a different error direction. Classifying the contact poses from the contact patterns is an image recognition problem. CNN has shown great success in image recognition in terms of efficiency and accuracy \cite{cnn}. We train one simple CNN to classify the contact patterns. The CNN has two convolutional layers, two pooling layers, and one fully-connected layer. The input data are the 3 most informative channels out of the 12-channel pattern. The output is the class of contact pose, which has 9 error directions for a square peg-hole and 11 error directions for a pentagonal peg-hole. Once the contact pose is identified, the peg will be guided towards the error direction with admittance control and inserted into the hole.

\subsection{Failure Recovery}
% From the exp, we observed failures cases even with system described above. reason .... how to solve ...
From the experiments, we observe failure cases even with the method described above. The reason is either the contact pose classification model predicts a wrong error direction or the admittance control fails to compensate small uncertainties. To increase the robustness of the proposed method, we add a failure recovery module. If we fail to insert the peg into the hole, the peg will be initialized to a slightly different pose than the original one, and redo the tilt-then-rotate strategy again.

%%%%%%%%%%%%%%%
\section{Simulations and experiments}

% In this section, we explain the simulations and experiments to validate the proposed method. 

\subsection{Simulations}

\subsubsection{Simulation Setup}
The simulated environment in MuJoCo is shown in Fig. \ref{fig:tilt_then_rotate}. The environment includes a peg and a hole, where the hole is fixed on the ground, and the peg is controlled by a well-tuned admittance controller. The side length of the hole is $50 mm$ and the side length of the peg is $49 mm$ (clearance = $1 mm$).  Contact force/torque is measured at the peg's center of mass.

\subsubsection{Data Collection}

A self-supervised scheme is applied to collect the data and build the contact pose mapping. As mentioned in \ref{assembly_strategy}, once the peg contacts the hole on the flat surface, the uncertainties in roll, pitch, and z-axis are eliminated. We only consider the remaining uncertainties in the x, y, and yaw axis. The contact poses are uniformly sampled from $x \in [-20, +20] mm$, $y \in [-20, +20] mm$, and $yaw \in [-3 \degree, +3 \degree]$. %The peg is always initialized at the origin. 
After the approaching stage in \ref{assembly_strategy}, the tilt-then-rotate strategy is applied, and $\alpha$ in equation (\ref{roll_pitch}) is set to $15 \degree$. The tilt-then-rotate motion is executed by the admittance controller in $N$ time steps, where $N= 2000$. The 12-dimension peg pose and contact force/torque are recorded in a matrix $A \in \mathbb{R}^{N \times 12}$. 
%We select 3 channels $A' \in R^{N \times 3}$ including position in $z$ axis $z$, torque in roll axis $M_x$, and torque in pitch axis $M_y$ to construct the contact patterns. 
The data of each dimension is normalized then smoothed by moving average with a window length $n = 20$, and the contact pattern is recorded in polar coordinates as a $12 \times 200 \times 200$ binary image. We label the contact patterns of a square peg-hole with 9 classes according to the initial contact poses (Fig. \ref{fig:classes}). The uncertainty in the yaw axis is compensated by the admittance controller and small oscillations in the yaw axis. 
%Considering large uncertainty in the yaw axis could be a potential future work. 
We also add $5\%$ noise to the parameters of the admittance controller in order to introduce variance to the collected data. We perform the self-supervised data collection for 5000 trails. The computation time is around 10 minutes. We split $80\%$ data as the training set and $20\%$ data as the test set.

\subsubsection{Model Training}

From the 12 channels contact patterns, we select 3 channels $A' \in \mathbb{R}^{N \times 3}$ including the position in $z$ axis $X_z$, the torque in roll axis $M_x$, and the torque in pitch axis $M_y$ as the input to the CNN. The reason that we select these 3 channels is that we experimentally find that these channels contain more features than other channels.  We downsample the contact patterns into $3 \times 20 \times 20$ images. We use an NVIDIA GeForce GTX 1080 Ti GPU for training. The training time is around 1 minute.

\subsubsection{Results}

The test accuracy of the contact pose classification neural network is $97.4\%$. Most of the failure cases are the contact pose at the boundary between two classes. We test on a second data set by collecting $1000$ data from a smaller square peg-hole, where the side length of the hole is $32 mm$ (clearance = $1 mm$). The test accuracy is $96.8\%$. This shows the generalization ability of the proposed method. Although the sizes of the parts, the contact measurements such as force, torque are different, the model still works very well. The reason is that we predict the contact pose according to the contact pattern, which is invariant to the size of the parts.

We perform another simulation experiment on a pentagonal peg-hole. The side length of the hole is $37 mm$ (clearance = $1 mm$). Because the contact pattern highly depends on the geometry of the peg-hole, we cannot apply the model learned from square peg-hole to pentagonal peg-hole. We redo the data collection and model training on the pentagonal pen-hole. Everything is the same as square peg-hole, except the number of contact pose classes becomes 11. The test accuracy is $91.0\%$.

We test the entire peg-in-hole assembly framework using the proposed method. We perform 100 trials on both square and pentagonal peg-hole. If the peg fails to be inserted into the hole, the failure recovery module will initialize the peg to a slightly different pose than the original one, and redo this trial again. If it requires more than 3 attempts to finish the task, we claim it fails. Table \ref{table1} shows the number of attempts needed to finish assembly in simulation. The high success rate shows that the proposed framework works well.

\begin{table}
\centering
\caption{Peg-in-hole assembly in simulation}
\label{table1}
\begin{tabular}{|l|l|l|l|l|l|l|l|l|}
\hline
\# of attempts  & 1 & 2  & 3 & $>3$ & total& success rate  \\ \hline
square ($50mm$) & 96 & 3  & 1 & 0 & 100 &$100\%$\\ \hline 
pentagon ($37mm$) & 82 & 11  & 3 & 4 & 100 &$96\%$\\ \hline
\end{tabular}
        
\end{table}

\subsection{Experiments}

\begin{figure*}[!ht]
	\centering
	\includegraphics[scale=0.525]{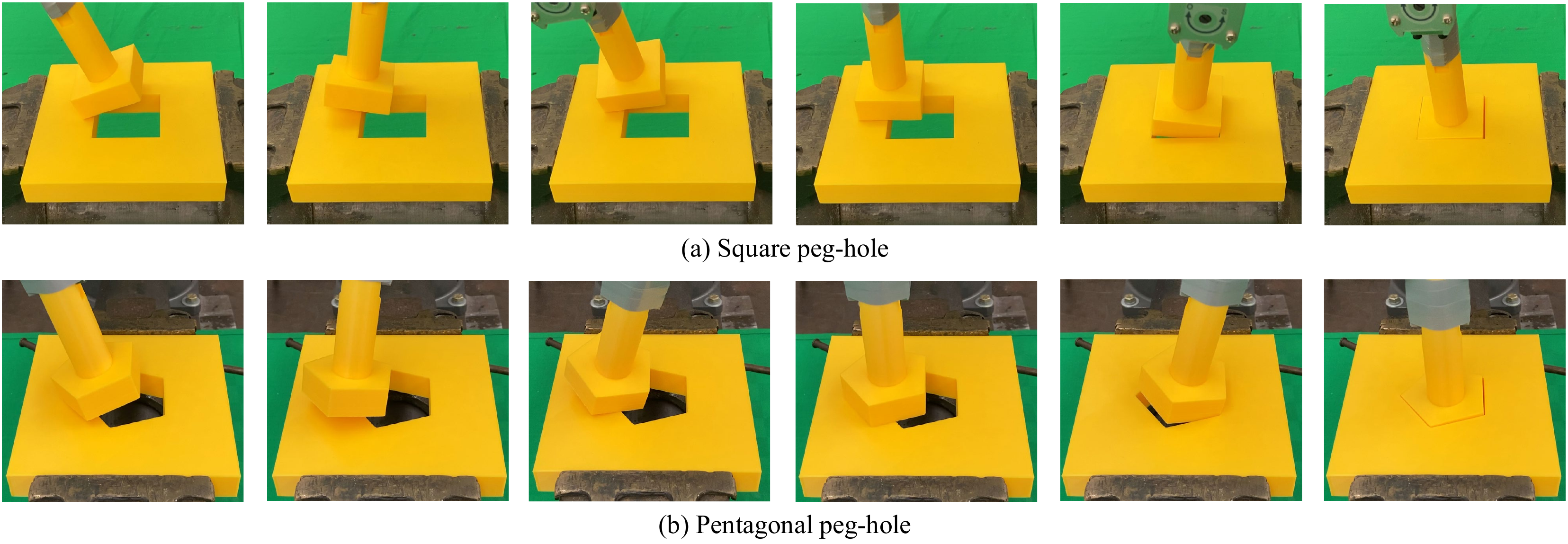}
	\caption{Snapshots of the experiments.}
	\label{fig:experiments}
\end{figure*}

\subsubsection{Experimental Setup}

The experiment environment (Fig. \ref{fig:experiments}) includes a 6 DOF FANUC LR-Mate 200iD, an ATI Mini45 F/T sensor, and 3D printed peg-holes. The F/T sensor is embedded in the robot end-effector to measure the force and torque during assembly. The force/torque measured at the robot wrist can be transfer to the force/torque at the peg's center of mass. The peg is fixed on the robot end-effector and the hole is fixed on a vise. The peg's pose can be controlled with an admittance controller at $125Hz$. The hole is randomly initialized with position and orientation uncertainties $\pm 20 mm$ and $\pm 3 \degree$, respectively. Three pairs of 3D printed peg-holes are tested, including a $50 mm$ square hole (clearance = $1mm$), a $32 mm$ square hole (clearance = $0.5mm$), and a $37 mm$ pentagonal hole (clearance = $1mm$). % The parameters are :?. 

\subsubsection{Results}

\begin{figure}
	\centering
	\includegraphics[scale=0.32]{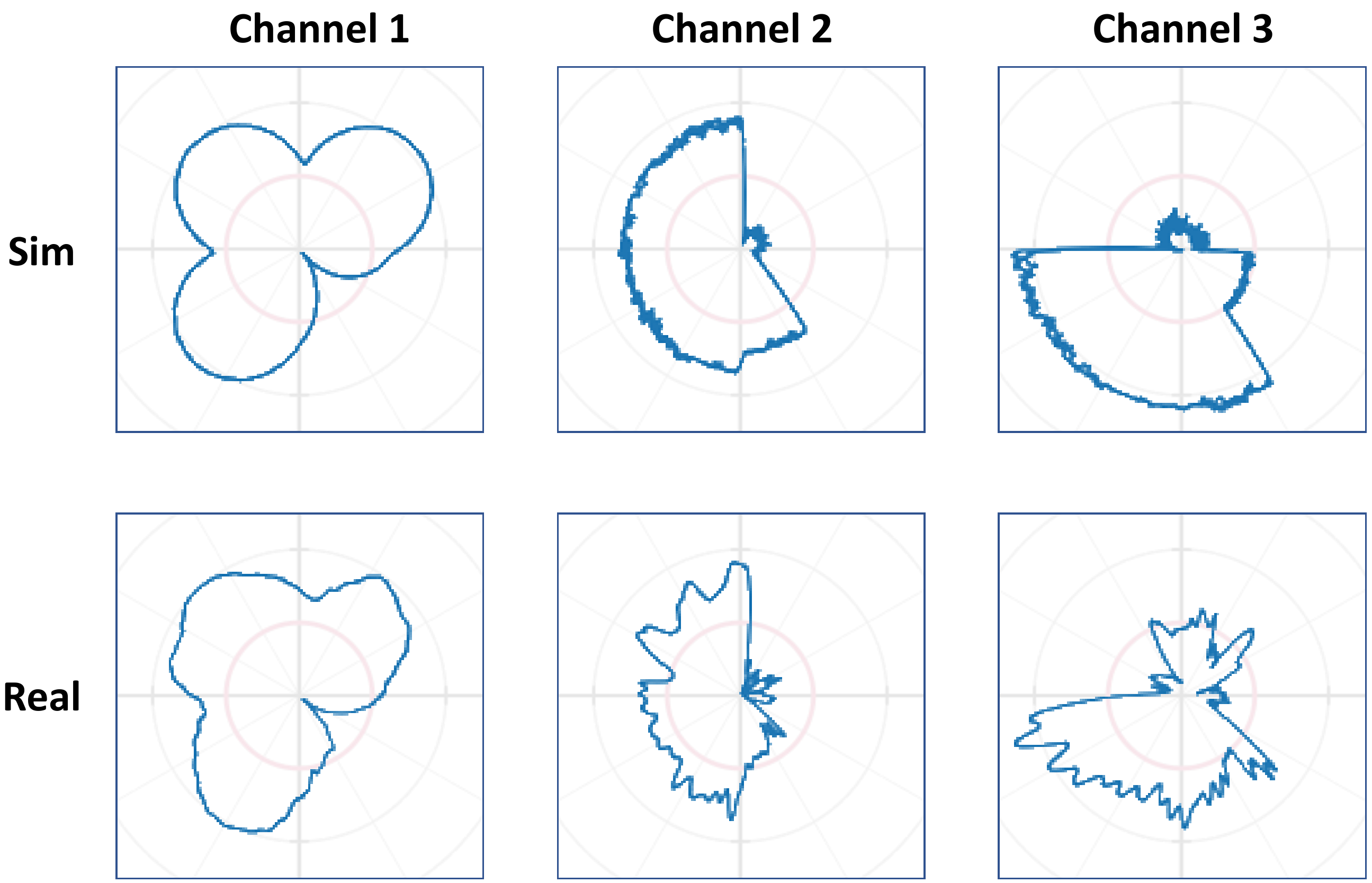}
	\caption{Comparison of the contact patterns in simulations and real-world experiments. They are generated from the same class of contact pose}
	\label{fig:sim2real}
\end{figure}

Fig. \ref{fig:sim2real} shows the comparison of the contact patterns generated from tilt-then-rotate strategy in simulation and real-world experiments. They are generated from the same class of contact pose. The data collected from the real-world has much noise than from simulation. Although there is a huge sim-to-real gap\cite{domain_random} between the simulated environment and the real world in terms of friction coefficient, inertia, stiffness, damping ratio, etc., we observe that the contact patterns do share similar features.

The contact pattern classification model learned in the simulation are applied to real-world experiments. Fig. \ref{fig:experiments} shows the snapshots of the assembly experiments. We perform 20 experiments on 3 different pairs of peg-hole respectively. Table \ref{table2} shows the number of attempts needed to finish assembly in real-world experiments. The model learned in simulation ($50mm$ square, clearance = $1mm$ ) can be successfully applied to real-world peg-hole of different sizes ($32mm$) and smaller clearance ($0.5mm$). This shows that the proposed method is able to tackle the sim-to-real gap. Supplementary videos can be found in \cite{website}.

\begin{table}
\centering
\caption{Peg-in-hole assembly in real-world experiments}
\label{table2}
\begin{tabular}{|l|l|l|l|l|l|l|l|l|}
\hline
\# of attempts & clearance & 1 & 2  & 3 & $>3$ & total & success rate\\ \hline
 square ($50 mm$)& $1 mm$  & 16 & 3  & 1 & 0  & 20 & $100\%$\\ \hline 
 square ($32 mm$)& $0.5 mm$  & 10 & 5  & 2 & 3  & 20 & $85\%$\\ \hline 
pentagon ($37 mm$) & $1mm$  & 15 & 3 & 1 & 1& 20 & $95\%$\\ \hline
\end{tabular}
        
\end{table}

\section{Discussion}

In this paper, we propose a novel framework to identify contact pose for peg-in-hole assembly under uncertainties. The proposed method utilizes a tilt-then-rotate strategy to generate contact patterns. A CNN is utilized to classify the contact poses %according to the contact patterns. The contact pose indicates the error direction which 
and guide the robot to achieve the assembly task with admittance control. Simulation and experiment results are provided to demonstrate the effectiveness of the proposed method. 
The main advantages of the proposed method include:
\begin{itemize}
 \item The injective mapping from the contact pattern to the contact pose.
 \item Robustness to sensor noise. %The force/torque sensor usually has a lot of noise in the contact-rich dynamic environment. Because the contact pattern is constructed using a sequence of contacts, which do not sensitive to one single measurement.
 \item The contact pose classification model is easy to obtain. All the training data can be quickly generated in simulation with a self-supervised scheme.
 \item Good generalization ability and small sim-to-real gap. Since the contact data is normalized and recorded in a polar coordinate, the pattern is sensitive neither to the size of the object nor the parameters of the admittance controller. A model learned from a larger peg-hole can be successfully applied to smaller ones as long as the geometries are the same. Furthermore, the model learned in simulation can be adapted to the real world, despite the huge sim-to-real gap.
\end{itemize}

    Here are the limitations of the proposed framework:
    \begin{itemize}
     \item The proposed method can only find the directions of the error, while it is unable to obtain the magnitude. In order to compensate for the error, the admittance controller needs to be well-tuned.
     \item The contact pose classification model can handle only position uncertainties, but it cannot classify the orientation uncertainties in the yaw axis.  
     %\item The contact pose classification model only works for one particular peg-hole shape. If the geometry of the peg-hole changes, we have to recollect the data and retrain the model.
    \end{itemize}
    
For future works, we plan to improve the algorithm so that it can handle orientation uncertainties and test it in more challenging scenarios. We also intend to incorporate active and adaptive sensing strategies to our framework. 

\bibliographystyle{IEEEtran}
\bibliography{bib}

% Generated by IEEEtran.bst, version: 1.14 (2015/08/26)
\begin{thebibliography}{10}
\providecommand{\url}[1]{#1}
\csname url@samestyle\endcsname
\providecommand{\newblock}{\relax}
\providecommand{\bibinfo}[2]{#2}
\providecommand{\BIBentrySTDinterwordspacing}{\spaceskip=0pt\relax}
\providecommand{\BIBentryALTinterwordstretchfactor}{4}
\providecommand{\BIBentryALTinterwordspacing}{\spaceskip=\fontdimen2\font plus
\BIBentryALTinterwordstretchfactor\fontdimen3\font minus
  \fontdimen4\font\relax}
\providecommand{\BIBforeignlanguage}[2]{{%
\expandafter\ifx\csname l@#1\endcsname\relax
\typeout{** WARNING: IEEEtran.bst: No hyphenation pattern has been}%
\typeout{** loaded for the language `#1'. Using the pattern for}%
\typeout{** the default language instead.}%
\else
\language=\csname l@#1\endcsname
\fi
#2}}
\providecommand{\BIBdecl}{\relax}
\BIBdecl

\bibitem{pose_estimate1}
Y.~Xiang, T.~Schmidt, V.~Narayanan, and D.~Fox, ``Posecnn: A convolutional
  neural network for 6d object pose estimation in cluttered scenes,'' in
  \emph{Robotics: Science and Systems (RSS)}, 2018.

\bibitem{pose_estimate2}
B.~Tekin, S.~N. Sinha, and P.~Fua, ``{Real-Time Seamless Single Shot 6D Object
  Pose Prediction},'' in \emph{CVPR}, 2018.

\bibitem{2001search}
S.~Chhatpar and M.~Branicky, ``Search strategies for peg-in-hole assemblies
  with position uncertainty,'' in \emph{2001 IEEE/RSJ International Conference
  on Intelligent Robots and Systems}, Maui, Hawaii, USA, 2001.

\bibitem{Drake1978UsingCI}
S.~Drake, ``Using compliance in lieu of sensory feedback for automatic
  assembly.'' 1978.

\bibitem{Whitney}
D.~E. Whitney, ``{Quasi-Static Assembly of Compliantly Supported Rigid
  Parts},'' \emph{Journal of Dynamic Systems, Measurement, and Control}, vol.
  104, no.~1, pp. 65--77, 03 1982.

\bibitem{handbook}
S.~Bruno and O.~Khatib, ``Springer handbook of robotics,'' 2008.

\bibitem{admittance}
C.~{Ott}, R.~{Mukherjee}, and Y.~{Nakamura}, ``Unified impedance and admittance
  control,'' in \emph{2010 IEEE International Conference on Robotics and
  Automation}, 2010, pp. 554--561.

\bibitem{2016threepoint}
T.~{Tang}, H.~{Lin}, {Yu Zhao}, {Wenjie Chen}, and M.~{Tomizuka}, ``Autonomous
  alignment of peg and hole by force/torque measurement for robotic assembly,''
  in \emph{2016 IEEE International Conference on Automation Science and
  Engineering (CASE)}, 2016, pp. 162--167.

\bibitem{2014rotate}
Y.~Kim, H.~Song, and J.~Song, ``\BIBforeignlanguage{English}{Hole detection
  algorithm for chamferless square peg-in-hole based on shape recognition using
  f/t sensor},'' \emph{\BIBforeignlanguage{English}{International Journal of
  Precision Engineering and Manufacturing}}, vol.~15, no.~3, pp. 425--432, Mar.
  2014.

\bibitem{2016teaching}
T.~{Tang}, H.~{Lin}, Y.~{Zhao}, Y.~{Fan}, W.~{Chen}, and M.~{Tomizuka}, ``Teach
  industrial robots peg-hole-insertion by human demonstration,'' in \emph{2016
  IEEE International Conference on Advanced Intelligent Mechatronics (AIM)},
  2016, pp. 488--494.

\bibitem{2016endtoend}
S.~Levine, C.~Finn, T.~Darrell, and P.~Abbeel, ``End-to-end training of deep
  visuomotor policies,'' in \emph{The Journal of Machine Learning Research},
  2016.

\bibitem{2019yongxiang}
Y.~{Fan}, J.~{Luo}, and M.~{Tomizuka}, ``A learning framework for high
  precision industrial assembly,'' in \emph{2019 International Conference on
  Robotics and Automation (ICRA)}, 2019, pp. 811--817.

\bibitem{2019syntheticdata}
J.~C. Triyonoputro, W.~Wan, and K.~Harada, ``Quickly inserting pegs into
  uncertain holes using multi-view images and deep network trained on synthetic
  data,'' \emph{CoRR}, vol. abs/1902.09157, 2019.

\bibitem{2019makingsense}
M.~A. {Lee}, Y.~{Zhu}, K.~{Srinivasan}, P.~{Shah}, S.~{Savarese}, L.~{Fei-Fei},
  A.~{Garg}, and J.~{Bohg}, ``Making sense of vision and touch: Self-supervised
  learning of multimodal representations for contact-rich tasks,'' in
  \emph{2019 International Conference on Robotics and Automation (ICRA)}, 2019,
  pp. 8943--8950.

\bibitem{2005particle}
S.~R. {Chhatpar} and M.~S. {Branicky}, ``Particle filtering for localization in
  robotic assemblies with position uncertainty,'' in \emph{2005 IEEE/RSJ
  International Conference on Intelligent Robots and Systems}, 2005, pp.
  3610--3617.

\bibitem{2007particle}
U.~{Thomas}, S.~{Molkenstruck}, R.~{Iser}, and F.~M. {Wahl}, ``Multi sensor
  fusion in robot assembly using particle filters,'' in \emph{Proceedings 2007
  IEEE International Conference on Robotics and Automation}, 2007, pp.
  3837--3843.

\bibitem{para_tunning}
L.~{Johannsmeier}, M.~{Gerchow}, and S.~{Haddadin}, ``A framework for robot
  manipulation: Skill formalism, meta learning and adaptive control,'' in
  \emph{2019 International Conference on Robotics and Automation (ICRA)}, 2019,
  pp. 5844--5850.

\bibitem{cnn}
A.~Krizhevsky, I.~Sutskever, and G.~E. Hinton, ``Imagenet classification with
  deep convolutional neural networks,'' in \emph{Advances in Neural Information
  Processing Systems 25}.\hskip 1em plus 0.5em minus 0.4em\relax Curran
  Associates, Inc., 2012, pp. 1097--1105.

\bibitem{domain_random}
J.~{Tobin}, R.~{Fong}, A.~{Ray}, J.~{Schneider}, W.~{Zaremba}, and P.~{Abbeel},
  ``Domain randomization for transferring deep neural networks from simulation
  to the real world,'' in \emph{2017 IEEE/RSJ International Conference on
  Intelligent Robots and Systems (IROS)}, 2017, pp. 23--30.

\bibitem{website}
{Supplementary videos of the tilt-then-rotate strategy.},
  {https://shiyujin0.github.io/TiltThenRotate/ACC2021.html}.

\end{thebibliography}

\vspace{12pt}

\end{document}